\documentclass[journal]{IEEEtran}

\usepackage{multirow}
\usepackage{enumerate}

\usepackage{cite}
\ifCLASSINFOpdf
  \usepackage[pdftex]{graphicx}
\else
  \usepackage[dvips]{graphicx}
  \DeclareGraphicsExtensions{.eps}
\fi
\usepackage[cmex10]{amsmath}
\usepackage{array}
\usepackage{mdwmath}
\usepackage{mdwtab}
\usepackage{eqparbox}
\usepackage[tight,footnotesize]{subfigure}
\usepackage[caption=true]{caption}
\usepackage[font=footnotesize]{subfig}
\usepackage{fixltx2e}
\usepackage{stfloats}
\usepackage{url}
\usepackage{multirow}
\usepackage{enumerate}
\usepackage{stfloats}
\usepackage{multicol}
\usepackage{algorithm}
\usepackage{amsmath}

\usepackage{algpseudocode}
\usepackage{amsmath}
\usepackage{graphics}
\usepackage{epsfig}
\usepackage{multirow}
\usepackage{multicol}

\usepackage{subfigure}
\usepackage{multirow}
\usepackage{enumerate}
\usepackage{stfloats}

\usepackage{color}

\usepackage{float}

\usepackage[font=footnotesize]{subfig}

\begin{document}

%
% paper title
% can use linebreaks \\ within to get better formatting as desired
%\title{HEp-2 Cell Classification using Discriminative Texture and Shape Features}
%\title{HEp-2 Cell Classification by Fusing Multiple Sources of Information}
\title{HEp-2 Cell Classification via Fusing Texture and Shape Information}

%\title{HEp-2 Cell Classification using Multi-Resolution Texture and Discriminative Shape Features}
%
%
% author names and IEEE memberships
% note positions of commas and nonbreaking spaces ( ~ ) LaTeX will not break
% a structure at a ~ so this keeps an author's name from being broken across
% two lines.
% use \thanks{} to gain access to the first footnote area
% a separate \thanks must be used for each paragraph as LaTeX2e's \thanks
% was not built to handle multiple paragraphs
%

\author{Xianbiao Qi, Guoying Zhao, ~\IEEEmembership{Senior Member,~IEEE}, Chun-Guang Li, ~\IEEEmembership{Member,~IEEE}, Jun Guo, \\ Matti Pietik{\"a}inen,~\IEEEmembership{Fellow,~IEEE}

%Xianbiao Qi, Guoying Zhao, ~\IEEEmembership{Senior Member,~IEEE,}, Chun-Guang Li, ~\IEEEmembership{Member,~IEEE,}, and \\Matti Pietik{\"a}inen,~\IEEEmembership{Fellow,~IEEE

%        Rong Xiao,
%        Yu Qiao, ~\IEEEmembership{Member,~IEEE}
%        Jun Guo,
%        %Lei Zhang, ~\IEEEmembership{Senior Member,~IEEE}
%        and Xiaoou Tang,~\IEEEmembership{Fellow,~IEEE}% <-this % stops a space

\thanks{Xianbiao Qi, Chun-Guang Li and Jun Guo are with the School of Information and Communication Engineering, Beijing University of Posts and Telecommunications, Beijing 100876, China. Xianbiao Qi is also with the Center of Machine Vision Research, University of Oulu, PO Box 4500, FIN-90014. E-mails:
qixianbiao@gmail.com, lichunguang@bupt.edu.cn, guojun@bupt.edu.cn.}
\thanks{Guoying Zhao and Matti Pietik{\"a}inen are with the Center of Machine Vision Research, University of Oulu, PO Box 4500, FIN-90014, Finland. E-mails: gyzhao@ee.oulu.fi, mkp@ee.oulu.fi.} }

\maketitle
%\boldmath
\begin{abstract}

Indirect Immunofluorescence (IIF) HEp-2 cell image is an effective evidence for diagnosis of autoimmune diseases. Recently computer-aided diagnosis of autoimmune diseases by IIF HEp-2 cell classification has attracted great attention. However the HEp-2 cell classification task is quite challenging due to large intra-class variation and small between-class variation.
In this paper we propose an effective and efficient approach for the automatic classification of IIF HEp-2 cell image by fusing multi-resolution texture information and richer shape information. To be specific, we propose to: a) capture the multi-resolution texture information by a novel Pairwise Rotation Invariant Co-occurrence of Local Gabor Binary Pattern (PRICoLGBP) descriptor, b) depict the richer shape information by using an Improved Fisher Vector (IFV) model with RootSIFT features which are sampled from large image patches in multiple scales, and c) combine them properly. We evaluate systematically the proposed approach on the IEEE International Conference on Pattern Recognition (ICPR) 2012, IEEE International Conference on Image Processing (ICIP) 2013 and ICPR 2014 contest data sets. The experimental results for the proposed methods significantly outperform the winners of
ICPR 2012 and ICIP 2013 contest, and achieve comparable performance with the winner of the newly released ICPR 2014 contest.

\end{abstract}

% IEEEtran.cls defaults to using nonbold math in the Abstract.
% This preserves the distinction between vectors and scalars. However,
% if the journal you are submitting to favors bold math in the abstract,
% then you can use LaTeX's standard command \boldmath at the very start
% of the abstract to achieve this. Many IEEE journals frown on math
% in the abstract anyway.

% Note that keywords are not normally used for peerreview papers.

\section{Introduction}
\label{sec:introduction}
Indirect immunofluorescence image (IIF) is an image analysis based diagnostic methodology to determine the existence of autoimmune diseases. Recently, it has attracted great attention due to its effectiveness.
More and more pattern recognition techniques \cite{foggia2013benchmarking, foggia2014PR, foggia2010early, percannella2012classification, theodorakopoulos2012hep, nosaka2013hep, theodorakopoulos2013hep, kong2013hep, shen2013hep, wiliem2013classification, han2014high, faraki2014fisher}  have been developed to make computer-aided diagnosis (CAD) of autoimmune diseases.
Before, manual labeling is the main approach for classifying the fluorescence patterns. However, the process of human labeling requires high expert knowledge, and meanwhile, it is also time consuming. Thus, to design a discriminative and robust HEp-2 cell classification system is extremely important.

\begin{figure}[t]
\begin{center}
\small
 \includegraphics[width=0.95\linewidth]{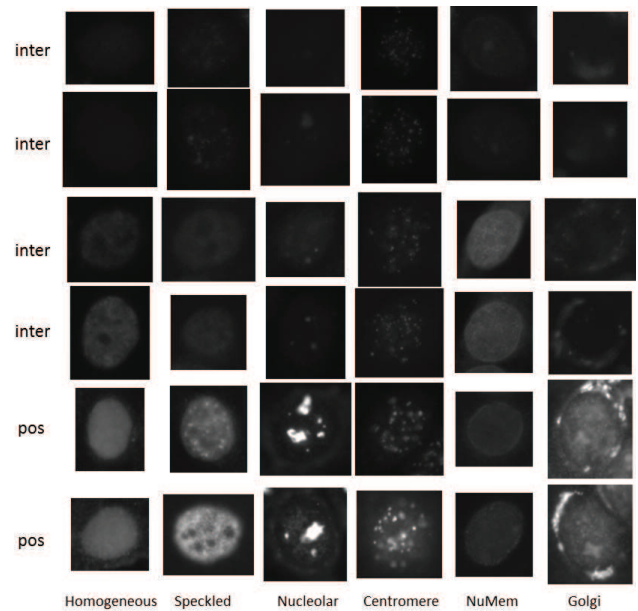}%{figures/examples.eps}
\end{center}
   \caption{Sample images from all six categories in ICIP 2013 contest data set with each column from one category. The six categories are ``Homogeneous'', ``Speckled'', ``Nucleolar'', ``Centromere'', ``NuMem'' and ``Golgi''. The first four rows show the ``intermediate'' samples, and the last two rows show the ``positive'' samples.}
\label{fig:icip}
\end{figure}

The HEp-2 cell classification task is challenging due to large intra-class and small between-class variations regardless of its importance. As shown in Fig. \ref{fig:icip}, the ``Intermediate'' and ``Positive'' cells from same categories have large variations, the ``Positive'' cells in raw images can be seen clearly, but the ``Intermediate'' cells can not be seen clearly. Meanwhile, some categories share similar shapes, such as the categories ``Homogeneous'' and ``Speckled'', and some categories show similar textures, such as the categories ``Nucleolar'' and ``Golgi''.

Recent ICPR 2012, ICIP 2013 and ICPR 2014 HEp-2 cell classification contests \cite{icpr2012hep2, icip2013, icpr2014hep2} have greatly put forward the development of HEp-2 cell analysis. Many features, image representation, classification methods were proposed or applied to this task. Currently, texture-based methods are the most widely used in this area. Local Binary Pattern (LBP) \cite{ojala1996comparative, ojala2002multiresolution, pietikainen2011computer} is widely recognized as a discriminative texture descriptor, and widely used in face recognition \cite{ahonen2006face}, static and dynamic texture classification \cite{ojala2002multiresolution, zhao2007dynamic}. Co-occurrence of adjacent LBP (CoALBP) \cite{nosaka2013hep}, Gradient-oriented Co-occurrence of LBPs (GoC-LBPs) \cite{theodorakopoulos2013hep} and pairwise rotation invariant co-occurrence of LBP (PRICoLBP) \cite{qi2014pairwise} are three of the best performing LBP variants in HEp-2 cell classification.  Besides of these three LBP variants, original LBP \cite{ojala2002multiresolution}, Completed LBP (CLBP) \cite{guo2010completed} were also used in the contests. Besides of LBP based texture features, some other famous texture features, such as Maximum Response Filter Banks (e.g. MR8) \cite{varma2005statistical}, Gray-Level Co-occurrence Matrices (GLCM) \cite{haralick1973textural}, Wavelet \cite{gabor1946theory}, were also used in this task. We also observed that Bag of Word \cite{csurka2004visual} model had been applied this task. %In BoW framework, several works focused on dictionary learning.

However, regardless of big improvement of classification accuracy in the past few years, previous works on HEp-2 cell classification task still have some limitations. Three key limitations are shown as follows:
\begin{itemize}

\item  Previous methods pay less attention to the multi-resolution texture information. Although texture information is widely studied, the influence of multi-resolution texture analysis to HEp-2 cell classification task is unknown.
\item Few works focus on capturing discriminative shape information. As far as we known, Vestergaard et al. \cite{Vestergaard2014} was the only work that explicitly explored the shape information in HEp-2 cell classification. Their work is different from the widely used Bag of Words (BOW) framework that our work is built on.
\item  The texture and shape information were considered individually, but they may be complementary to each other in practice. Thus, it will be interesting to investigate their complementary properties between them.

\end{itemize}

In this work, we attempt to address the pending issues mentioned above and hence our contributions are highlighted as follows:
\begin{itemize}
 \item  We explore the effect of multi-resolution texture for HEp-2 cell classification. To be specific, we capture the multi-resolution texture information by a novel Pairwise Rotation Invariant Co-occurrence of Local Gabor Binary Pattern (PRICoLGBP) descriptor, which is able to capture multi-resolution texture information effectively.
 \item We propose an effective method to depict the richer shape information by using an Improved Fisher Vector (IFV) model with RootSIFT features. Different from previous work, we extract local features from large image patches in multiple scales.
 \item We investigate the complementary effect of texture and shape information. By combining the multi-resolution texture and richer shape information, we yield superior classification accuracy. Compared with the winner of ICPR 2012 contest, our methods improves the accuracy of the winner by about 7\%. Compared with the winner of ICIP 2013 contest, our method obtains 4\% higher accuracy. Our method also achieves comparable performance to the winner of the newly release ICPR 2014 contest.
\end{itemize}

The rest of the paper is organized as follows. We firstly review the state-of-the-art methods in the HEp-2 cell classification area in Sec. \ref{sec:relatedworks}. Then, we present the proposed texture and shape features in detail in Sec. \ref{sec:mainbody}. The used data sets are introduced in Sec. \ref{sec:dataset}. In Sec. \ref{sec:experiments}, we firstly
 give a comprehensive experimental evaluations of properties of the proposed discriminative texture and shape methods, and then compare it with some state-of-the-art methods. Finally, we give a conclusion in Sec. \ref{sec:conclusion}.

\section{Related Works}
\label{sec:relatedworks}
%\textcolor{blue}{In this section, we will firstly introduce the methods of the $1$-st and $2$-nd places in ICPR 2012 contest, then describe the approaches of $1$-st and $2$-nd places in ICIP 2013 contest, and then introduce two well-performing works that are highly relevant to this paper, and finally give some remarks about existing HEp-2 cell classification methods.}

\subsection{Best Performing Methods in ICPR 2012 Contest}

{\bf{Nosaka}} et al. \cite{nosaka2013hep}-the winner of ICPR 2012-only used the green channel in their method. The image was filtered by a Gaussian function to remove the noise. To improve the robustness to image rotation, they manually rotated the image to 9 orientations. Then, they extracted co-occurrence of adjacent LBP (CoALBP) features for all images (including the original images and the manually created images). Finally, they trained a linear Support Vector Machine (SVM) classifier.

  The success of Nosaka's methods is due to the following three aspects:
\begin{itemize}
 \item  Strong discriminative of CoALBP: the CoALBP was built on LBP that proves to be a powerful texture descriptor. Moreover, to capture strong spatial layout information, the CoALBP used 10 templates.

 \item  Green channel used: Among all the three channels, green channel was much stronger than the red and blue channels. Using gray-scale image would weaken the texture information in the green channel.

 \item  Manually creating many rotated training samples:  To improve the robustness of CoALBP to image rotation, they manually rotate the imaged to 9 orientations, and created 9 new rotated training samples.

\end{itemize}
  Regardless of its success on ICPR 2012 contest, this method also has some limitations. Firstly, since the CoALBP itself is not rotation invariant, thus, the CoALBP is not robust to image rotation although Nosaka et al. try to improve the CoALBP's robustness to rotation by manually creating more rotated training samples. Secondly, the discriminative power of CoALBP is limited due to that the CoALBP is built on the co-occurrence of two LBPs with four neighbors. The LBP(4, 1) is usually considered to be less discriminative than the LBP(8, 1).

{\bf{Kong}} et al. \cite{kong2013hep}-the second place of ICPR 2012- adopt Varma's MR8 method to extract the texture feature. The local regions were normalized before the filter responses are applied. After feature extraction, they trained a global texton dictionary using K-means clustering. Thus, each image could be represented as a frequency histogram of textons. They also used a pyramid histogram of oriented gradients (PHOG) \cite{bosch2007representing} feature to depict the shape information. The texture and shape histogram were concatenated with different weights. Finally, they used a K-Nearest Neighbor (KNN) classifier with ${\chi}^2$ distance.

%We believe the following two aspects may greatly affect the final classification accuracy. First, the used shape and texture feature may not be discriminative enough. Second, it has been proved in many applications, KNN classifier always shows worse performance compared with Support Vector Machine (SVM) classifier.

\subsection{Best Performing Methods in ICIP 2013 Contest}

{\bf{Shen}} et al. \cite{icip2013}-the winner of ICIP 2013- combined the the original PRICoLBP and the Bag of SIFT feature. For the PRICoLBP feature, they used 10 templates. The dimension of the PRICoLBP for each template is 590. Thus, the total dimension of their used PRICoLBP\footnote{http://qixianbiao.github.io/} feature is 5900. For the Bag of SIFT feature, following the traditional bag of words model, they created 1024 words using K-means clustering. Finally, they concatenated these two features and used linear SVM (Support Vector Machine) with square root features.

The success of this method is due to the following three aspects. Firstly, the PRICoLBP is good at capturing the texture information, meanwhile, as argued in \cite{qi2014pairwise}, when the shape structures are
strong in the data set, the utilization of 10 templates significantly improves the performance of 2 templates. Secondly, the bag of SIFT is good at capturing the global texture and shape information. Finally, the square root normalization of the feature is an
effective method for linear SVM. The square root normalization has proved to be effective in many computer vision works \cite{vedaldi11efficient}.

{\bf{Vestergaard}} et al. \cite{Vestergaard2014}-the merit winner of ICIP 2013- adopted a standard pipeline for the supervised image classification: preprocessing of the images, feature extraction and classification. A two-stage preprocessing method was exploited. First, each image $I$ was augmented with its logarithmic representation $loge(I)$. Then, the logarithmic representation was mapped linearly to [0,1]. For the feature extraction, Vestergaard et al. extracted three kinds of features including: 1) the ``Intersity'' of each image (Negative/Intermediate/Positive) as an integer flat, 2) morphological features extracted from the provided mask (containing the area of the mask region, eccentricity, major and minor axis length, perimeter); and 3) the donut-like shape index histogram feature (for both image representations). For the classification, Vestergaard et al. used a RBF kernel SVM.

\subsection{Best Performing Methods in ICPR 2014 Contest}
{\bf{Manivannan}} et al. \cite{manivannan2014hep} ranked 1st in the newly released ICPR 2014 HEp-2 cell classification contest \cite{icpr2014hep2}. Their method can be summarized into the following steps:
\begin{itemize}
\item[a)] Rotating the images to four orientations (0, 90, 180, 270) respectively;
\item[b)] Dense sampling of multi-scale patches ($12 \times 12, 16 \times 16, 20 \times 20$);
\item[c)] Extraction of four types of features (Multi-resolution local patterns (mLP), Root-SIFT (rSIFT), Random projections (RP), Intensity histogram (IH));
\item[d)] Feature encoding with Locality-constrained Linear Coding (LLC) for four types of features and four orientations individually. Thus, $4\times 4 = 16$ histograms can be obtained;
\item[e)] Training 16 classifiers with linear SVM and Classification based on 16 classifiers.

\end{itemize}

\subsection{Other Relevant and Well-Performing Methods}
%\textcolor{blue}{In this subsection, we introduce several approaches that are highly relevant with our methods. These mentioned methods below did not show in above-mentioned contest, but demonstrates good performance on HEp-2 cell classification task.}

{\bf{Theodorakopoulos}} et al. \cite{theodorakopoulos2013hep} proposed a sparse representation of textural features which were fused into dissimilarity space. Along with a multivariate distribution of SIFT feature, Theodorakopoulos et al. \cite{theodorakopoulos2013hep} proposed a Gradient-oriented Co-occurrence of LBPs which is considered in \cite{theodorakopoulos2013hep} as a relaxed variation of the PRICoLBP. The descriptors were fused while creating a dissimilarity representation of an image. Finally, a sparse representation-based classification scheme was used for the classification.

In \cite{theodorakopoulos2013hep}, the usage of SIFT feature was in a simple manner. Simple multivariate distribution of SIFT feature was used. Meanwhile, the used GoC-LBP was not robust to image rotation. Since the GoC-LBP was built on the co-occurrence of two uniform LBPs,
its dimension ($59\times 59 = 3481$) was higher than PRICoLBP (590).

{\bf{Faraki}} et al. \cite{faraki2014fisher} extended the traditional bag-of-word (BOW) from Euclidean space to non-Euclidean Riemanian manifolds that is an intrinsic bag of Riemannian words (BoRW). The BOW model has been applied to HEp-2 cell in \cite{wiliem2013classification} before.
Faraki et al. also proposed Fisher Tensor to encode higher statistics information when building the histogram for the images. The Fisher Tensor can be seen as a Riemannian version of Fisher Vector \cite{perronnin2010improving}. Their proposed BoRW and its extension with Fisher Tensor in \cite{faraki2014fisher} demonstrate great performance on both HEp-2 cell classification and texture classification tasks.

\section{Hep-2 Cell Classification Using Discriminative Texture and Shape Features}
\label{sec:mainbody}
This section consists of three subsections. In the first part, we introduce one novel multi-resolution texture feature. In the second part, we present our approach for depicting discriminative shape information. Finally, we describe the normalization and classification methods.

\subsection{Discriminative Texture Feature}

\subsubsection{\bf{Local Binary Pattern}}

Local Binary Pattern (LBP) that was firstly proposed by Ojala et al. \cite{ojala2002multiresolution} is considered as a simple and effective texture descriptor.
For any pixel $A$ in an image, we can compute its LBP pattern by thresholding the pixel values of its circularly symmetric
$P$ neighbors with the pixel value of the central point $A$. The LBP of pixel $A$ can be defined as
follows:
\begin{equation}\nonumber
    \text{LBP}_{P,R}(A) = \sum_{i=0}^{P-1} s(V_i - V_c)2^i, \ \
    s(x) = \begin{cases}
    1, &  x \ge 0 \\
    0, &  x < 0, \\
    \end{cases}
\end{equation}
where $P$ is the number of the neighbors, $R$ is the radius, $V_c$ is the pixel value of point $A$, and $V_i$ is the pixel value of point $A$'s $i$th neighbor. Since the $s(V_i-V_c)$ is invariant to monotonic change of illumination, thus the LBP is gray-scale invariant.

The patterns with very few spatial transitions is considered to depict the fundamental image micro-structures. Such patterns were called as ``uniform patterns''.
Ojala et al. \cite{ojala2002multiresolution} defined a uniformity measure for the uniform patterns, which is $\mathcal{U}(\text{LBP}_{P,R}(A)) \le N$ ($N$ is usually set to 2). The uniformity measure $\mathcal{U}(\cdot)$ can be calculated as follows:
\begin{equation}\nonumber
\mathcal{U}(\text{LBP}_{P,R}(A)) = \sum_{i=0}^{P-1}|s(V_i - V_c) - s(V_{i-1}-V_c)|,
\end{equation}
where the pixel value of $V_{-1}$ is equivalent to the pixel value of $V_{P-1}$. For example, ``11000000" and ``10000001" are uniform patterns, and ``10000100" and ``10101100" are non-uniform patterns.

Rotation invariant LBP ({${\text{LBP}^{ri}}$}) and rotation invariant uniform LBP ({${\text{LBP}^{ru}}$})
are also introduced in \cite{ojala2002multiresolution}. The ${\text{LBP}^{ri}}$ can be defined as:

\begin{equation}\nonumber
    \text{LBP}_{P, R}^{ri}(A) = \text{min}\{\text{ROR}(\text{LBP}_{P, R}(A), i) \ \  | \ \  \forall i\in[0,P-1]\},
    \label{eq:riLBPCode}
\end{equation}
where $ROR(x , i)$  performs a circularly bit-wise right shift for $i$ times. The $\text{LBP}_{n, r}^{ru}(A)$ is defined as
\begin{equation}\nonumber
\label{eq:rui}
    \text{LBP}_{P, R}^{ru}(A) = \begin{cases}
    \sum_{i=0}^{P-1} s(V_i-V_c), &  \mathcal{U}(\text{LBP}(A)) \le 2 \\
    P+1, &  \text{otherwise}, \\
    \end{cases}
\end{equation}

The $\text{LBP}_{8,1}$ has 256 patterns in total, in which 58 patterns are uniform and the rest 198 patterns are non-uniform. Usually, the 198 non-uniform patterns are summarized to one pattern. Thus, 59 patterns are usually used for uniform LBP. The rotation invariant uniform $\text{LBP}_{8,1}$ includes 10 patterns.

%Above, we have introduced the original LBP and its uniform and rotation invariant variants.

\begin{figure}[h]
\begin{center}
\small
 \includegraphics[width=1.0\linewidth]{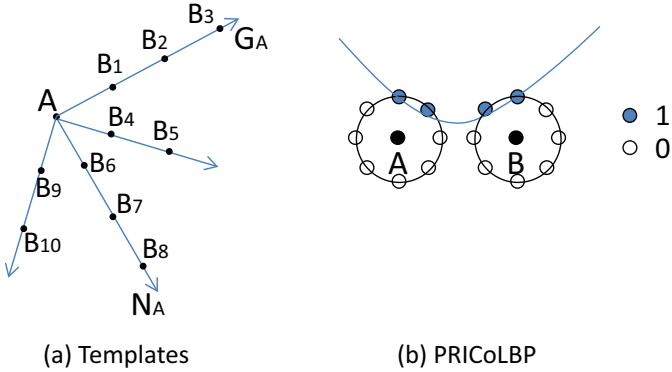}%{figures/examples.eps}
\end{center}
   \caption{Pairwise Rotation Invariant Co-occurrence of Local Binary Pattern. The left of the figure shows 10 templates based on the reference point $A$. $G_A$ is the gradient orientation of point $A$, $N_A$ is the normal direction of point $A$. The right point shows one co-occurrence pair of LBPs. PRICoLBP captures stronger structures than the single LBP.}
\label{fig:pricolbp}
\end{figure}

\subsubsection{\bf{Single-Resolution Texture Information}}
Pairwise rotation invariant co-occurrence LBPs (PRICoLBP) is recently introduced by Qi et al. \cite{qi2014pairwise} for texture related tasks. As shown in Fig. \ref{fig:pricolbp}(a), the PRICoLBP is built on the two adjacent LBP points. Given a point $A$, the PRICoLBP contains the following two key steps to calculate its rotation invariant pattern:

\begin{itemize}
\item According to the gradient and normal orientation (Normal orientation is the direction that is orthogonal to the gradient orientation.) of point $A$ and pre-defined templates as shown in Fig. \ref{fig:pricolbp}(a), the position of point $B_i$ can be uniquely determined. The gradient orientation can be calculated as $\theta(A) = \arctan(\frac{dy(A)}{dx(A)})$.
\item With a pair $A$ and $B_i$, pairwise rotation invariant encoding was used to encode the co-occurrence of two LBPs.
\end{itemize}
In practice, we used the gradient magnitudes of point $A$ and $B_i$ to weight their co-pattern.

For the first step, given a point $A$, the PRICoLBP uses the following equation to determine the position of point $B_i$:
\begin{equation}
\label{eq:MSMOCoLBP}
\setlength{\abovedisplayskip}{0pt}
\setlength{\belowdisplayskip}{1pt}
B_i  = a_i*G(A) + b_i*N(A) + A,
\end{equation}
where $a_i$ and $b_i$ are pre-defined coefficients for template $i$, and $G(A)$ and $N(A)$ are the gradient and normal directions of point $A$. In practice, we can choose 10 pairs for [$a_i$, $b_i$] as shown in Fig. \ref{fig:pricolbp}(a), one pair corresponds to one template.

\begin{figure}[t]
\begin{center}
\small
 \includegraphics[width=1.0\linewidth]{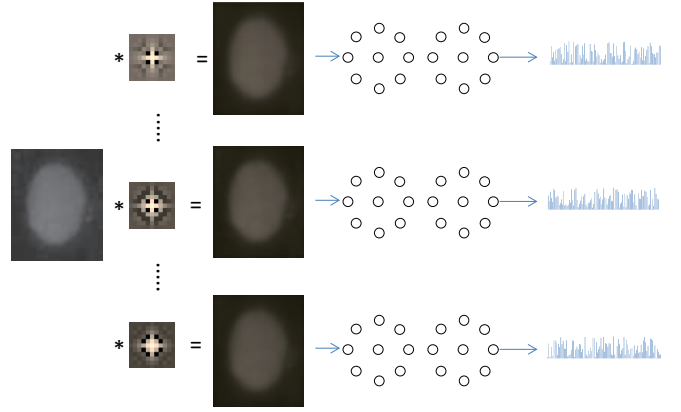}%{figures/examples.eps}
\end{center}
   \caption{The framework of PRICoLGBP. First, the image is filtered by Gabor filters with different scales. Then, the PRICoLBP features are extracted from each filtered images. Finally, the features extracted from each scale will be concatenated into the final feature.}
\label{fig:gaborcolbp}
\end{figure}

When the point pair $A$ and $B_i$ are determined, a pairwise rotation invariant encoding strategy is used to encode the pair. Denote $\text{LBP}^{u}(A,i)$ as the uniform LBP of point $A$ by using $i$-th index as the start point of the binary sequence. The PRICoLBP can be defined as follows:
\begin{equation}
\label{eq:PriCoLBP}
\text{PRICoLBP}_{P, R}(A,B_i)=[\text{LBP}_{P, R}^{ru}(A),\text{LBP}_{P, R}^{u}(B_i, i(A))]_{co},
\end{equation}
where $i(A)\in \{0,P-1\}$ is an index, which can be determined by minimizing the binary sequence of point $A$. $[\ ,\ ]_{co}$ is a co-occurrence operator firstly introduced in \cite{haralick1973textural}. Suppose $\text{LBP}_{P, R}^{ru}(A)$ has $M$ patterns, and $\text{LBP}_{P, R}^{u}(B_i, i(A))$ has $N$ patterns, then their co-occurrence has $M\times N$ patterns.

For one pair $A$ and $B_i$ with $P = 8$ and $R = 1$, $\text{LBP}_{8, 1}^{ru}(A)$ has 10 patterns, $\text{LBP}_{8, 1}^{u}$ has 59 patterns,
thus, the dimension of $\text{PRICoLBP}_{8, 1}$ is $10\times 59 = 590$. If 10 templates are used as shown in Fig. \ref{fig:pricolbp}(a), the dimension for PRICoLBP is $10\times 590 = 5900$.

\subsubsection{\bf{Multi-Resolution Texture Information}}

\begin{multicols}{2}
\begin{figure*}[t]
\begin{center}
\small
 \includegraphics[width=1.0\linewidth]{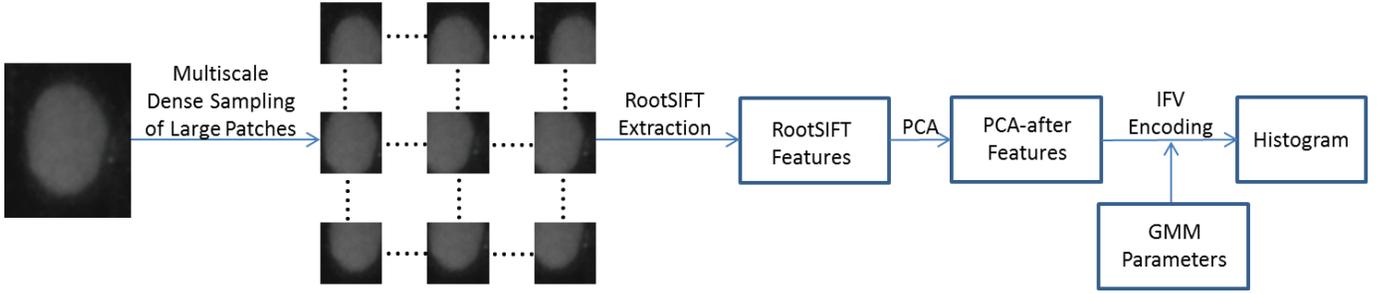}%{figures/examples.eps}
\end{center}
   \caption{The framework of discriminative shape information description. First, we densely sample thousands of large patches in multiple scales with a preset step (such as 2), then the RootSIFT feature is extracted for each patch. PCA is applied to the RootSIFT. Finally, IFV encoding is applied to the PCA-after features with pre-learned GMM parameters. The obtained histogram from IFV encoding can be input into a linear SVM for the final classification.}
\label{fig:Framework}
\end{figure*}
\end{multicols}

 The PRICoLBP is effective to capture the structures in the small scales
  (such as co-occurrence of $\text{LBP}_{8,1}$ and  co-occurrence of $\text{LBP}_(8,2)$), but texture information in large scales is ignored. However, multi-resolution texture
  information is always effective for many vision applications.

To capture multi-resolution texture information, we propose a novel pairwise rotation invariant co-occurrence of local Gabor binary pattern (PRICoLGBP) descriptor. Gabor wavelet \cite{gabor1946theory} \cite{movellan2002tutorial} is an effective filter to
capture multi-resolution and multi-orientation information. The PRICoLGBP is built on the Gabor filter and PRICoLBP descriptor. The framework of our PRICoLGBP can be seen in Fig. \ref{fig:gaborcolbp}. We convolute the original
image with different Gabor filters, and then extract the PRICoLBP from each filtered image, and finally
concatenate all PRICoLBPs into the final feature.
In experiments, we found that the PRICoLGBP is not sensitive to rotation variation for the Gabor filtered
images, thus, we only use one pre-fixed orientation for all scales.

The PRICoLGBP shares some similar properties with Local Gabor Binary
Pattern (LGBP) \cite{zhang2005local} that is seen as a powerful LBP variants in face recognition, but
different from the LGBP, our PRICoLGBP is built on a more discriminative co-occurrence of LBPs features.
Thus, we can expect that PRICoLGBP can capture stronger multi-resolution texture information.

We believe two strong properties of the PRICoLGBP makes it effective for
IIF HEp-2 cell classification.

\begin{itemize}

\item PRICoLGBP has strong texture discrimination. In IIF HEp-2 cell classification, texture-based
methods proves to be effective.

\item Gabor and PRICoLBP both are robust to image illumination variation. PRICoLGBP inherited the
properties from both Gabor and PRICoLBP. In IIF HEp-2 cells, the ``Positive'' and ``Intermediate'' cells
from the same categories show extremely varying illumination.
\end{itemize}

\subsection{Effective Shape Feature}
In this subsection, we present an effective method to depict the richer shape information by using an Improved Fisher Vector (IFV) model with RootSIFT features extracted from large image patches in multiple scales. Our approach consists of three steps: a) patch sampling, b) feature description with RootSIFT, and c) encoding by IFV. The flowchart to illustrate our approach is displayed in Fig.~\ref{fig:Framework}.

%\begin{itemize}
%\item Patch Sampling to obtain the most representative and discriminative regions.
%\item Feature Description to effectively capture the structures of the sampled regions.
%\item Encoding Methods to preserve the strong discrimination of the features.
%\end{itemize}

%In this section, we will mainly introduce our approach to depict discriminative shape information.
%Fig. \ref{fig:Framework} has illustrated the framework of our approach. In the following, we will introduce three key steps in the framework: patch sampling, feature description and encoding methods. To fully capture the shape information, these three steps are all indispensable:
%\begin{itemize}
%\item Patch Sampling to obtain the most representative and discriminative regions.
%\item Feature Description to effectively capture the structures of the sampled regions.
%\item Encoding Methods to preserve the strong discrimination of the features.
%\end{itemize}

% \noindent
%\subsubsection{\bf{Discriminative Patch Sampling for depicting shape information.}}
\subsubsection{\bf{Patch Sampling for Depicting Shape Information}}
To increase the discriminativeness in shape information, we propose to sample large patches, since that the large patches preserve stronger shape structures. To be specific, instead of sampling patches of small size, e.g., $16\times 16$, $17\times 17$, or $19\times 19$ as in object categorization tasks, we sample much larger patches, e.g., $41\times 41$. We can observe in Fig. \ref{fig:Framework} that the sampled patches cover more than $\frac{1}{3}$ of the whole image.\footnote{In general, a HEp-2 cell image is of $65\times 70$ and hence preserve stronger shape structure from the sampled patches. % texture information dominates the local regions (such as $3\times 3$ and $5\times 5$ regions), and shape information becomes stronger along with the increase of the patch size.
}
%Our motivation to focus on capturing shape information is based on the observation that shape information is rich and discriminative for both ``positive'' and ``intermediate'' HEp-2 cells.

\begin{figure}[h]
\begin{center}
\small
 \includegraphics[width=1.0\linewidth]{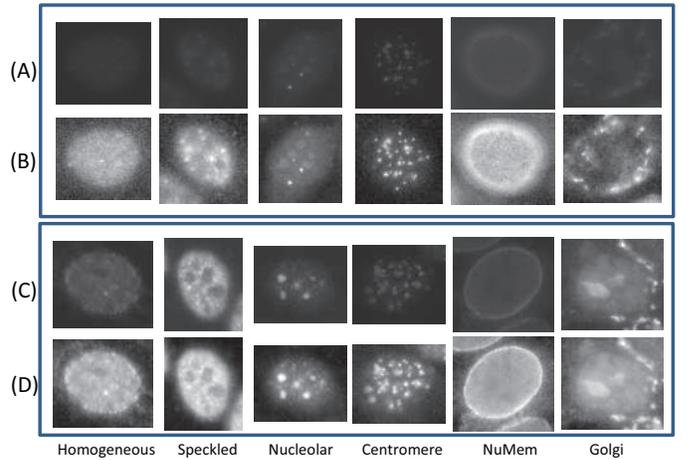}%{figures/examples.eps}
\end{center}
   \caption{Illustration of discrimination of shape information in ICIP 2013 contest data set. Each column comes from one category. It is easy to find that the shapes vary a lot among different categories. (A) shows the ``Intermediate'' cell images from each category, and (B) shows (A)'s corresponding enhanced images from a logarithmic operator $loge(I)$. (C) shows the ``Positive'' cell images, and (D) is (C)'s corresponding enhanced images.}
\label{fig:enhancement}
\end{figure}

In Fig. \ref{fig:enhancement}, we show some samples from all six categories in ICIP 2013 contest
data set.\footnote{To visualize the shape structures clearly, we enhance the images at first by using a logarithmic operator $loge(I)$ on the image and then normalize the image to the range of $[0,255]$. This preprocessing method was proposed in \cite{Vestergaard2014}.
Notice that:}
\begin{itemize}
\item The shape structures from different categories change a lot. Each category has its own basic characteristics. For instance, the category ``NuMum'' has bright and thick boundary, the category ``Centermere'' has many bright spots, and the category ``Golgi'' does not have well-formed boundary. Considering local texture structures, the shape difference between some categories is large. For instance, the categories ``Nucleolar'' and ``Centromere'' are easy to differentiate when jointly considering the shape and texture.
\item The ``positive'' and ``intermediate'' HEp-2 cells from same category share similar shape structure, %although sometimes,    it is not \textcolor{blue}{easy for humans}
    although we cannot see the shape structure of the ``intermediate'' HEp-2 cells clearly.
\end{itemize}
These observations are the rationales to explore the shape information for HEp-2 cell image classification.
%Based on the above-mentioned two observations, we can conclude that shape information is discriminative for HEp-2 cell images.

%To capture the discriminative shape information, we propose to sample large patches because we
%believe that large patches can preserve strong shape structures. Instead of sampling
%patches like $17\times 17$ pixels or $19\times 19$ pixels as usual object recognition tasks, we sample
%larger patches (e.g. $41\times 41$).
%The sampled patches can be seen in Fig. \ref{fig:Framework}. They cover more than $\frac{1}{3}$ of
%the whole image (e.g. $65\times 70$). We can clearly see its shape structure from the sampled patches, which means that the sampled patches preserves the strong shape information. We believe that texture
%information dominates the local regions (such as $3\times 3$ and $5\times 5$ regions), and
%shape information becomes stronger along with the increase of the patch size.

\subsubsection{\bf{RootSIFT Feature Extraction on Large Patches}}
We extract 128-dimensional SIFT features \cite{lowe2004distinctive} from the sampled large patches. For each SIFT feature $\bf f$, we normalize it with $L_1$-norm and then take the componentwise square root operation, i.e.,
\begin{align}
\begin{split}
\bar {\bf f} &= {\bf f \over \| \bf f \|_1}, \\
\tilde f_j   &= \sqrt{\bar f_j},~~~~j=1,\cdots, 128.
\end{split}
\end{align}
The obtained $\tilde f$ is termed as ``RootSIFT'' \cite{arandjelovic2012three}, which was proposed by Relja et al. to enhance the discriminative power of SIFT.

\subsubsection{\bf{Improved Fisher Vector (IFV) for Encoding the RootSIFTs}}
We encode the RootSIFT features by Improved Fisher Vector (IFV) approach \cite{perronnin2010improving} \cite{sanchez2013image}, which consists of three steps:
\begin{itemize}
\item Data decorrelation by Principal Component Analysis (PCA).
\item Training a Gaussian Model of Mixture (GMM).
\item Forming the IFV by using the first and second order statistics in GMM.
\end{itemize}

Denote the parameters in GMM as $\{ \pi_k, \mu_k, \Sigma_k\}_{k=1}^K$ where $\pi_k$ is the membership probability, $\mu_k$ is the mean of $k$-th component Gaussian, and $\Sigma_k$ is the covariance matrix which is enforced to be diagonal.
Let $X=[\textbf{x}_1, \textbf{x}_2, ..., \textbf{x}_N]\in \mathcal{R}^{d\times{N}}$ be a set of feature vectors of an image after decorrelation, where $d$ is reduced feature dimension of using PCA and $N$ is the number of RootSIFT features in the image. IFV captures the deviation of the features in an image from the first and second statistics of the GMM. %IFV measures the first and second order variation between the features and the GMM components. Given a set of features $X = [x_1,...,x_N]$, we can calculate $s_{ki}$, i.e. the probability of feature $x_i$ belonging to $k$-th Gaussian component.
To be specific, IFV is defined as follows:
\begin{equation}
{\bf{F}} = [{{\bf{u}}_1}^{\rm{T}},{\bf{v}}_1^{\rm{T}},..., {\bf{u}_K}^{\rm{T}},{\bf{v}_K}^{\rm{T}}]^{\rm{T}},
\end{equation}
where
\begin{equation}
 {\bf{u}}_k = \frac{1}{N \sqrt{{\pi}_k}} \sum_{i=1}^{N}s_{ki} {{\Sigma}_k}^{-\frac{1}{2}} (x_i - \mu_k),
\end{equation}
\begin{equation}
 {\bf{v}}_k = \frac{1}{N \sqrt{2{\pi}_k}} \sum_{i=1}^{N}s_{ki} [(x_i - \mu_k) \cdot ({{\Sigma}_k}^{-1} (x_i - \mu_k) ) - {\bf{1}}],
\end{equation}
in which $s_{ki}$ is defined as
\begin{equation}\nonumber
 s_{ki} =  \frac{ p(x_i|\mu_k,{\Sigma}_k) {\pi}_k}{ \sum_{i=1}^{K} p(x_i|\mu_i,{\Sigma}_k)){\pi}_i }, \ \ \ \ \ \ k=1,...,K .
\label{eq:gmmS}
\end{equation}
The parameter $s_{ki}$ is the responsibility of feature $\textbf{x}_i$ belonging to the $k$-th component in the GMM.

Note that the dimension of ${\bf{F}}$ is $2\times{d}\times{K}$.% $d$ is the dimension of our RootSIFT after PCA.
In our experiments, we set $d$ as 80, and $K$ as 256, the number of mixture components in GMM. The final dimension of IFV feature is  $2\times 80 \times 256 = 41960$. Note also that this is the first time that IFV is used in HEp-2 cell classification task.

%\subsubsection{\bf{Summary:}} To capture discriminative shape information, we have the following three key contributions  in this subsection,
%\begin{itemize}
%\item We propose to sample large patches to capture stronger shape information. The sampled
%patches cover more than $\frac{1}{3}$ of the whole image.
%\item We propose to use the RootSIFT feature to depict the sampled patches. We believe
%this spatial division in RootSIFT can depict shape information well.
%\item We propose to use IFV to encode the RootSIFT to preserve the discrimination of RootSIFT feature.
%
%%This is motivated by the recent success of the IFV in object recognition.
%
%\end{itemize}

\subsection{\bf{Histogram Normalization and Classification}}
Histogram normalization is a key step before training a SVM model.
We normalize the histogram componentwisely as follows:
\begin{equation}
\label{eq:chiB}
    %H_i = sign(H_i) \sqrt{\mid{H_i}\mid}, \forall \ i \in [1, D],
\hat H_i = sign(H_i) \sqrt{\mid{H_i}\mid}, \forall i=1,\cdots,D,
\end{equation}
where $D$ is the dimension of $H$, $sign(\cdot)$ is a sign function. And then we further normalize the histogram  with $L_2$ norm.

For classification we use \textit{linear} SVM since it is widely used in large scale problems. For linear SVM, the training is fast and the speed of classification in test phase is also fast, compared to kernel SVM. We use the \textit{one-vs-the-rest} strategy to handle the multi-class classification problem.

\begin{figure}[h]
\begin{center}
\small
 \includegraphics[width=1.0\linewidth]{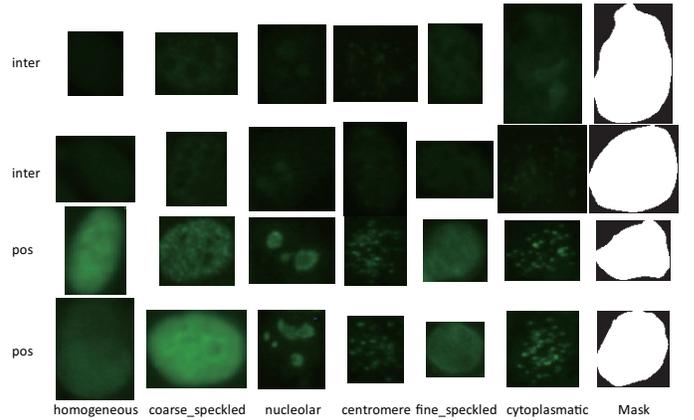}%{figures/examples.eps}
\end{center}
   \caption{Sample images from ICPR 2012 contest dataset with each column from one category. The first two rows show the intermediate samples, and the last two rows show the positive samples. All the images are shown in their original image size.}
\label{fig:icpr}
\end{figure}

\section{Datasets and Evaluation Metrics}
\label{sec:dataset}

%
%
%
%In this section, we will firstly introduce the ICPR 2012 contest data set, and then introduce the ICIP 2013 data set, and finally briefly introduce the evaluation metrics used in this paper.
%

\begin{table}[h]
\caption{Details of ICPR 2012 contest data set.}              % title of Table
\centering                                     % used for centering table
%\large
\normalsize
\begin{tabular}{| @{\ }c@{\ } | @{\ }c@{\ } | @{\ }c@{\ } |@{\ }c@{\ } | @{\ }c@{\ } | @{\ }c@{\ } | @{\ }c@{\ } | @{\ }c@{\ } |}                       % centered columns (4 columns)
\hline
                 &   Homo    &    Coar     &    Fine    &    Nucl    &    Cent    &   Cyto   & Total   \\[0.05cm]
\hline
Instances/train  &   3    &    2     &    2    &    2    &    3    &   2        &  14     \\
\hline
  Cells/train          &   150  &    109   &    94  &    102  &    208  &   60      &  723  \\
\hline
Instances/test   &   2    &    3     &    2    &    2    &    3    &   2        &  14     \\
\hline
  Cells/test          &   180  &    101   &    114  &    139  &    149  &   51      &  734  \\
\hline
\end{tabular}
\label{table:icprdataset}                   % is used to refer this table in the text
\end{table}

\subsection{ICPR 2012 Contest Dataset}
ICPR 2012 cell images were acquired by means of a fluorescence microscope (40-fold magnification) coupled with
a 50W mercury vapor lamp and with a digital camera.
The images have a resolution of $1388\times 1038$ pixels, a
color depth of 24 bits and they are stored in an uncompressed format.
Specialists manually segmented and annotated each
cell. In particular, a biomedical engineer manually segmented the cells by the use of a tablet PC. Subsequently,
each image was verified and annotated by a medical
doctor specialized in immunology.
The dataset contains 28 images almost equally distributed with respect to the different patterns.
In the contest, the 28 images are divided into training and testing sets. The information for training and testing sets is shown in Tab. \ref{table:icprdataset}. More detailed information can be found in \cite{icpr2012hep2}.
Some samples are shown in Fig. \ref{fig:icpr}.

Note that a specimen always has dozens of cells. The cells in the same specimen always have higher similarity than that of the cells from different specimens. Thus, to evaluate the generalization ability of the methods, the cells in one specimen can only be used for training or testing, it will be misleading to split them into training and testing. In the ICPR 2012 contest report, several methods used this strategy and directly splits all cell images instead of the specimens into training and validation sets, but their final results reported by the organizers were significantly lower than the authors' reported results.

%
%
%The ICPR dataset consists of six categories:
%
%
%\begin{itemize}
%
% \item {\bf{Centromere:}} characterized by serveral discrete speckles distributed throughout the interphase nucler and characteristically found in the condensed nuclear chromatin during mitosis as a bar of closely associated speckles;
%
% \item {\bf{Nucleolar:}} characterized by clustered large granules in the nucleoli of interphase cells which tend towards homogeneity, with less than six granules per cell;
%
% \item {\bf{Homogeneous:}} characterized by a diffuse staining of the interphase nuclei and staining of the chromatin of mitotic cells;
%
% \item {\bf{Fine Speckled:}} characterized by a fine granular nuclear staining of the interphase cell nuclei;
%
% \item {\bf{Coarse Speckled:}} characterized by a coarse granular nuclear staining of the interphase cell nuclei;
%
% \item {\bf{Cytoplasmic:}} characterized by fine fluorescent fibers running over the length of the cell.
%
%\end{itemize}
%The ICPR 2012 contest data set are color images. Sample images from ICPR 2012 are shown in Fig. \ref{fig:icpr}. For each image in the data set, a mask image is also provided. In Tab. \ref{table:icprdataset}, we present some details of ICPR 2012 contest data set.

\subsection{ICIP 2013 Contest Dataset}
%ICIP 2013 contest dataset was built between 2011 and 2013 at Sullivan Nicolaides Pathlogy laboratory, Austrilia. The dataset utilises 419 patient positive sera which were
%prepared on the 18-well slide of HEP-2000 IIF assay from
%Immuno Concepts N.A. Ltd. with screening dilution 1:80.
%The specimens were then automatically photographed using
%a monochrome high dynamic range cooled microscopy camera which was fitted on a microscope with a plan-Apochromat
%20x/0.8 objective lens and an LED illumination source. Approximately 100-200 cell images were extracted from each
%patient serum.
%%
%%In total there were 68,429 cell images extracted. We divided these into 13,596 images for training and
%%54,833 for testing.

The ICIP 2013 data set uses 419 patients positive sera with screening dilution 1:80. The specimens were automatically captured using a monochrome high dynamic range cooled microscopy camera. For each patient serum, 100-200 cell images were extracted. In total, there were 68429 cell images extracted. The whole 68429 cell images were divided into 13596 training samples and 54833 testing samples.

\begin{table}[h]
\caption{Details of ICIP 2013 contest data set.}              % title of Table
\centering                                     % used for centering table
%\large
\normalsize
\begin{tabular}{| @{\ }c@{\ } | @{\ }c@{\ } | @{\ }c@{\ } |@{\ }c@{\ } | @{\ }c@{\ } | @{\ }c@{\ } | @{\ }c@{\ } | @{\ }c@{\ } |}                       % centered columns (4 columns)
\hline
               &   Ho    &    Sp     &    Nu    &    Ce    &    NM    &   Go   & Total   \\[0.05cm]
\hline
Specimens      &   16    &    16     &    16    &    16    &    15    &   4        &  83     \\
\hline
%Positive       &   16    &    16     &    16    &    16    &    15    &   4        &  83     \\
%\hline
%Intermediate   &   16    &    16     &    16    &    16    &    15    &   4        &  83     \\
%\hline
  Cells        &   2494  &    2831   &    2598  &    2741  &    2208  &   724      &  13596  \\
\hline
\end{tabular}
\label{table:icipdataset}                   % is used to refer this table in the text
\end{table}

The labeling process involved at least two scientists who
read each patient's specimen under a microscope. A third expert's opinion was sought to adjudicate any discrepancy between the two opinions. In this way, a ground-truth mask can be extracted from each cell image.
%We used each specimen label for the
%ground-truth of cells extracted from it. Furthermore, all the
%labels were validated by using secondary tests such as ENA,
%and anti-ds-DNA to confirm the presence and/absence of specific patterns

%
%
%\begin{itemize}
%
% \item {\bf{Homogeneous:}} a uniform diffuse fluorescence covering the entire nucleoplasm sometimes accentuated in the nuclear periphery;
%
% \item {\bf{Speckled:}} these patterns have two categories:
%
%    - coarse speckled: densely distributed, variously sized speckles, generally associated with larger speckles, throughout nucleoplasm of interphase cells; nucleoli are negative;
%
%    - fine speckled: fine speckled staining in a uniform distribution, sometimes very dense so that an almost homogeneous pattern is attained; nucleoli may be positive or negative;
%
% \item {\bf{Nucleolar:}} brightly clustered large granules corresponding to decoration of the fibrillar centers of the nucleoli as well as the coiled bodies;
%
% \item {\bf{Centromere:}} rather uniform discrete speckles located throughout the entire nucleus;
%
% \item {\bf{Golgi:}} staining of a polar organelle adjecent to and partly surrounding the nucleus, composed of irregular large granules. Nuclei and nucleoli are negative. Diffuse staining of the cytoplasm of dividing cells sometimes with accentuation around chromosomal material;
%
% \item {\bf{Nuclear Membrane:}} a smooth homogeneous ring-like fluorescence of the nuclear membrane in interphase cells.
%
%\end{itemize}

The testing images are not released. But the training set is big enough to evaluate different algorithms. Some basic information for the training data in ICIP 2013 contest are shown in Tab. \ref{table:icipdataset}. More detailed information can be found in \cite{icip2013}. Some sample images are shown in Fig. \ref{fig:icip}.

It should be noted that in ICPR 2014 contest, the Task-1 used the same dataset as ICIP 2013 contest.

\subsection{Evaluation Metrics}
In the previous ICPR 2012 and ICIP 2013 contests, accuracy of maximum classification number is used as a performance metric. For specimen, in ICPR 2012 data set, the testing number of images are 734, if the 500 images are classified correctly, then the accuracy is $\frac{500}{734}$. In this paper, we follow the metric of the previous ICPR 2012 and ICIP 2013 contest, and use the maximum classification number as the metric.

When comparing our method with ICPR 2014 winner \cite{manivannan2014hep}, we strictly follow the winner's protocol, and use the leave-one-specimen-out protocol. The averaged Mean Class Accuracy (MCA) is reported.
%Mean class accuracy (MCA) is another widely used performance metric. The difference between them comes from the unbalance of the testing samples in each category.

\section{Experiments}
\label{sec:experiments}

%
%
%In this experimental section, we will firstly introduce some implementation details, then evaluate some properties of the proposed methods, and finally compared our methods with the state-of-the-art methods on ICPR 2012 and ICIP 2013  contest data sets individually.

\subsection{Implementation Details}
{\bf{PRICoLGBP.}} For multi-resolution PRICoLGBP feature, we use the original image and 7 Gabor-filtered images under 7 different scales $\{1, 2, 3, 4, 5, 6, 7\}$. For each filtered image, we can extract one PRICoLBP feature. In each PRICoLBP feature, we use 10 templates. As we described before, the dimension of PRICoLBP using one template is 590. Thus, the final dimension for PRICoLGBP is $8\times10\times590=47200$.

{\bf{RootSIFT(IFV).}} We densely sample the RootSIFT feature at six scales with grid step 2. The sampled patch size is $41\times 41$. If the image size (height or width) is less 64, we will resize it to the image with minimum size 64 and keep the height/width ratio.
Six scales are achieved by filtering the images with Gaussians with different scales of different standard deviates $\{{1.5}, {1.5}^2, {1.5}^3, {1.5}^4, {1.5}^5, {1.5}^6\}$.
For specimen, for an image with image size $70\times 70$, we can sample 225 points for each scale. Thus, for six scales, we can get 1350 sampled patches. For a larger image, such as $70\times 80$, we will sample more points.
In the IFV, we firstly sample 100000 RootSIFT features from the training samples, then the
100000 RootSIFT features are used to learn the PCA components, and 80 principal components are preserved as the basis for dimension reduction. As pointed out by \cite{sanchez2013image}, the PCA is a key step in the IFV framework. With above-mentioned 100000 after-PCA RootSIFT feature, we learn a Gaussian Mixture Model (GMM) with 256 components. For the PCA, we use the built-in SVD (Singular Value Decomposition). For the GMM, we use Vlfeat to learn the parameters $\theta = \left \{ \pi_k, \mu_k, \Sigma_k, k=1,\cdots,K\right \}$.
The final dimension using the IFV encoding is $2\times80\times256 = 41960$.

{\bf{Experimental Setups.}} Vlfeat toolbox \cite{vedaldi2010vlfeat} is used for fast RootSIFT extraction and IFV encoding, and Liblinear \cite{fan2008liblinear} is used for the linear SVM training and classification. For the parameter C, we cross-validated it in \{0.001, 0.01, 0.1, 1, 100, 1000\}. It should be noted that the first author of this paper provides PRICoLBP feature and classifier for Shen et al. (the ICIP 2013 winner). We share the source code that had been submitted into ICIP 2013 and achieved the 1st place.
All experimental comparisons are conducted in the same framework. Take ICIP 2013 contest data set as example, first, we create 10 splits for 10 repeated experiments. For each split, the whole ICIP contest 2013 data set are randomly divided into the training and testing sets. Meanwhile, to truly show the generalization performance of approaches, the images from the same cell are only divided into training or testing set. Thus, All comparisons are fair in this paper. We have provided the matlab code\footnote{\url{https://www.dropbox.com/s/eoifdhqjs1o7vky/HEp2Cell.zip?dl=0}} to repeat the experimental results.
%For the regularization $C$ in the linear SVM, we fix it as 40 for all experiments.
%We also have the source code for ICIP winner's algorithm.

\subsection{Evaluation of Features}
In this subsection, we will mainly evaluate some aspects of the proposed texture and shape features. The ICPR 2012 data set is too small to fully evaluate the properties of the proposed methods. Thus, we will use ICIP 2013 data set in this subsection. To fully evaluate the properties, we use four sets of different experimental setups, as shown in Tab. \ref{table:icipsetup}.
Take the setup ``D'' as an example, in experimental setup ``D'', 42 specimens (including 8 specimens from ``Homogeneous'', 8 specimens from ``Speckled'', 8 specimens from ``Nucleolar'', 8 specimens from ``Centromere'', 8 specimens from ``NuMem'' and 2 specimens from ``Golgi'') in all 83 specimens are used for training, and the rest 41 specimens are used for testing, each specimen includes 100-200 cell images. Using this strategy, the images in one specimen can only be divided into training or testing. This used strategy can truly reflect the generalization ability because the images come from the specimen usually have higher similarity than that between images from different specimen, if part of the images in one specimen are used for training, the rest images that are used for testing are easily correctly classified, but this strategy can not be generalized to other unknown specimen. We pre-create 10 training and testing splits randomly. We repeat the experiments 10 times and average the results.

\begin{table}[h]
\caption{Four experimental setups for ICIP 2013 data set. For example, ``1'' means all images from one specimen are used for training, and the remaining specimens are used for testing. }              % title of Table
\centering                                     % used for centering table
%\large
\normalsize
\begin{tabular}{| c | c | c | c | c | c | c | c |}                       % centered columns (4 columns)
\hline
           &   Ho    &    Sp     &    Nu    &    Ce    &    NM    &   Go   & total   \\[0.05cm]
\hline
  Setup A        &   1     &    1      &    1     &    1     &    1     &   1        &  6     \\
\hline
  Setup B        &   2     &    2      &    2     &    2     &    2     &   2        &  12  \\
\hline
  Setup C        &   4     &    4      &    4     &    4     &    4     &   2        &  22  \\
\hline
  Setup D        &   8     &    8      &    8     &    8     &    8     &   2        &  42  \\
\hline
\end{tabular}
\label{table:icipsetup}                   % is used to refer this table in the text
\end{table}

{\bf{Evaluation of Multi-Resolution Texture Extraction Strategy.}} Here, we conduct experiments to compare the PRICoLBP and PRICoLGBP on above-mentioned four experimental setups. The results are shown in Tab. \ref{table:MultiResolution}.

\begin{table}[h]
\caption{Comparison between the single-resolution and multi-resolution texture feature on ICIP 2013 data set.}              % title of Table
\centering                                     % used for centering table
%\large
%\normalsize
\small
\begin{tabular}{| @{}c@{} | @{\ }c@{\ } | @{\ }c@{\ } | @{\ }c@{\ } | @{\ }c@{\ } | }
\hline
                 &   Setup A       &   Setup B   &  Setup C     &  Setup D         \\[0.1cm]
\hline
PRICoLBP         &   $43.1 \pm{6.8}$  &    $55.2 \pm{5.4}$ &    $67.6 \pm{4.0}$  &    $73.9 \pm{2.9}$ \\ [0.1cm]
\hline
PRICoLGBP   &   $50.4 \pm{6.2}$  &    $61.4 \pm{5.1}$ &    $72.2 \pm{3.8}$  &    $78.4 \pm{2.7}$ \\ [0.1cm]
\hline
\end{tabular}
\label{table:MultiResolution}                   % is used to refer this table in the text
\end{table}

We can observe that from Tab. \ref{table:MultiResolution}, multi-resolution texture feature significantly improves the single-resolution texture feature. For specimen, the multi-resolution PRICoLGBP improves the PRICoLBP by 7.3\% and 4.5\% for the experimental setup ``A'' and ``D''.
%It should be noted that in Shen, the combination between PRICoLBP (73.9\%) and Bag-of-SIFT with VQ (73.6\%) improves the performance to 75.1\%.

{\bf{Evaluation of Improved Fisher Vector Encoding.}} To evaluate the effectiveness of the Improved Fisher Vector, we compare it with the traditional Vector Quantization (VQ). For both VQ and IFV, the feature is normalized according to Eq. \ref{eq:chiB}. A linear SVM is used for training and classification. The results averaged on 10 random repeats are shown in Tab. \ref{table:IFV}.

\begin{table}[h]
\caption{Comparison between Vector Quantization and Improved Fisher Vector on ICIP 2013 data set.}              % title of Table
\centering                                     % used for centering table
%\large
%\normalsize
\small
\begin{tabular}{| @{\ }c@{\ } | @{\ }c@{\ } | @{\ }c@{\ } | @{\ }c@{\ } | @{\ }c@{\ } | }
\hline
                 &   Setup A       &   Setup B   &  Setup C     &  Setup D        \\[0.1cm]
\hline
RootSIFT(VQ)        &   $54.9 \pm{6.8}$  &    $62.2 \pm{4.5}$ &    $68.3\pm{2.4}$  &    $71.2 \pm{2.3}$ \\ [0.1cm]
\hline
RootSIFT(IFV)         &   $58.7 \pm{6.7}$  &    $66.7 \pm{5.1}$ &    $74.7 \pm{2.7}$ &    $78.4 \pm{2.7}$ \\ [0.1cm]
\hline
\end{tabular}
\label{table:IFV}                   % is used to refer this table in the text
\end{table}

 From Tab. \ref{table:IFV}, we can find that the IFV encoding method sharply improves the performance of the VQ encoding method. For specimen, under the experimental configuration ``D'', the IFV improves the VQ from 71.2\% to 78.4\%. In conclusion, the IFV is an effective way to preserve the discriminative power of the features under the BoW framework.

%
%{\bf{Evaluation of the Patch Size.}}
%
%
%\begin{table}[h]
%\caption{Performance evaluation of FV encoding with SIFT under different patch size .}              % title of Table
%\centering                                     % used for centering table
%\large
%%\normalsize
%\begin{tabular}{| c | c | c | c | c |  }
%\hline
%                 &   A       &    B      &    C     &    D     \\[0.1cm]
%\hline
%$17 \times 17$   &   60.10   &    60.10  &    60.10 &    60.10 \\[0.1cm]
%\hline
%$21 \times 21$   &   70.10   &    70.10  &    70.10 &    70.10 \\[0.1cm]
%\hline
%$27 \times 27$   &   70.10   &    70.10  &    70.10 &    70.10 \\[0.1cm]
%\hline
%$35 \times 35$   &   70.10   &    70.10  &    70.10 &    70.10 \\[0.1cm]
%\hline
%$41 \times 41$   &   70.10   &    70.10  &    70.10 &    70.10 \\[0.1cm]
%\hline
%\end{tabular}
%\label{table:MultiResolution}                   % is used to refer this table in the text
%\end{table}
%
%We could find that large patch always show better performance than small patch. This may reflect that the shape information is effective on IIF cell classification. Large patch can better capture shape information.

{\bf{Evaluation of Normalization Method.}} Here, we evaluate the importance of the normalization method. For both PRICoLGBP feature and RootSIFT(IFV), we normalized the histograms according to Eq. \ref{eq:chiB}. We compare them with the direct $L_2$ normalized histograms(without using Eq. \ref{eq:chiB}) under the linear SVM framework.
The results averaged on 10 random repeats are shown in Tab. \ref{table:Normalization}.

%
%\begin{multicols}{2}
%\begin{table*}[h]
%\caption{ICIP 2013.}              % title of Table
%\centering                                     % used for centering table
%%\large
%\normalsize
%\begin{tabular}{| c | c | c | c |}                       % centered columns (4 columns)
%\hline
%           &   PRICoLBP   &    BoW of SIFT &    PRICoLBP + BoW of SIFT     \\[0.05cm]
%\hline
%Accuracy  &    20.0       &    20.0       &          20.0        \\[0.05cm]
%\hline
%           &   Gabor PRICoLBP   &    FV with RootSIFT &    Gabor PRICoLBP   +    FV with RootSIFT    \\[0.05cm]
%\hline
%Accuracy  &    20.0       &    20.0       &          20.0        \\[0.05cm]
%\hline
%\end{tabular}
%\label{table:icipstate}                   % is used to refer this table in the text
%\end{table*}
%\end{multicols}

\begin{table}[h]
\caption{Evaluation of Normalization Methods on RootSIFT(IFV) and PRICoLGBP on ICIP 2013 data set. In the table, ``PRICoLGBP'' and ``RootSIFT(IFV)'' denotes the feature under the direct $L_2$ normalization, ``PRICoLGBP*'' and ``RootSIFT(IFV*)'' denotes the features normalized by Eq. \ref{eq:chiB} and then by $L_2$ normalization.}              % title of Table
\centering                                     % used for centering table
%\large
%\normalsize
\small
\begin{tabular}{| @{\ }c@{\ } | @{\ }c@{\ } | @{\ }c@{\ } | @{\ }c@{\ } | @{\ }c@{\ } |  }
\hline
                 &   Setup A       &   Setup B   &  Setup C     &  Setup D         \\[0.1cm]
\hline
PRICoLGBP           &   $48.4 \pm{6.2}$   &    $59.4 \pm{4.6}$  &   $70.3 \pm{3.2}$ &    $77.5 \pm{3.2}$ \\[0.1cm]
\hline
PRICoLGBP*          &   $50.4 \pm{6.2}$   &    $61.4 \pm{5.1}$ &    $72.2 \pm{3.8}$ &    $78.4 \pm{2.7}$ \\ [0.1cm]
\hline
RootSIFT(IFV)             &   $57.8 \pm{7.0}$   &    $65.7 \pm{5.4}$ &    $73.6 \pm{2.9}$ &    $77.4 \pm{2.7}$ \\[0.1cm]
\hline
RootSIFT(IFV*)            &   $58.7 \pm{6.7}$   &    $66.7 \pm{5.1}$ &    $74.7 \pm{2.7}$ &    $78.4 \pm{2.7}$ \\[0.1cm]
\hline
\end{tabular}
\label{table:Normalization}                   % is used to refer this table in the text
\end{table}

From Tab. \ref{table:Normalization}, it is easy to find that the PRICoLGBP with normalization according to Eq. \ref{eq:chiB} consistently outperforms the PRICoLGBP without normalization, and the RootSIFT(IFV) using normalization also consistently outperforms the non-normalized feature.
In conclusion, the normalization always improves the classification accuracy.
%we can find that the used normalization method consistently boosts the classification performance.

\begin{multicols}{2}
\begin{figure*}
\begin{center}
\small
 \includegraphics[width=0.92\linewidth]{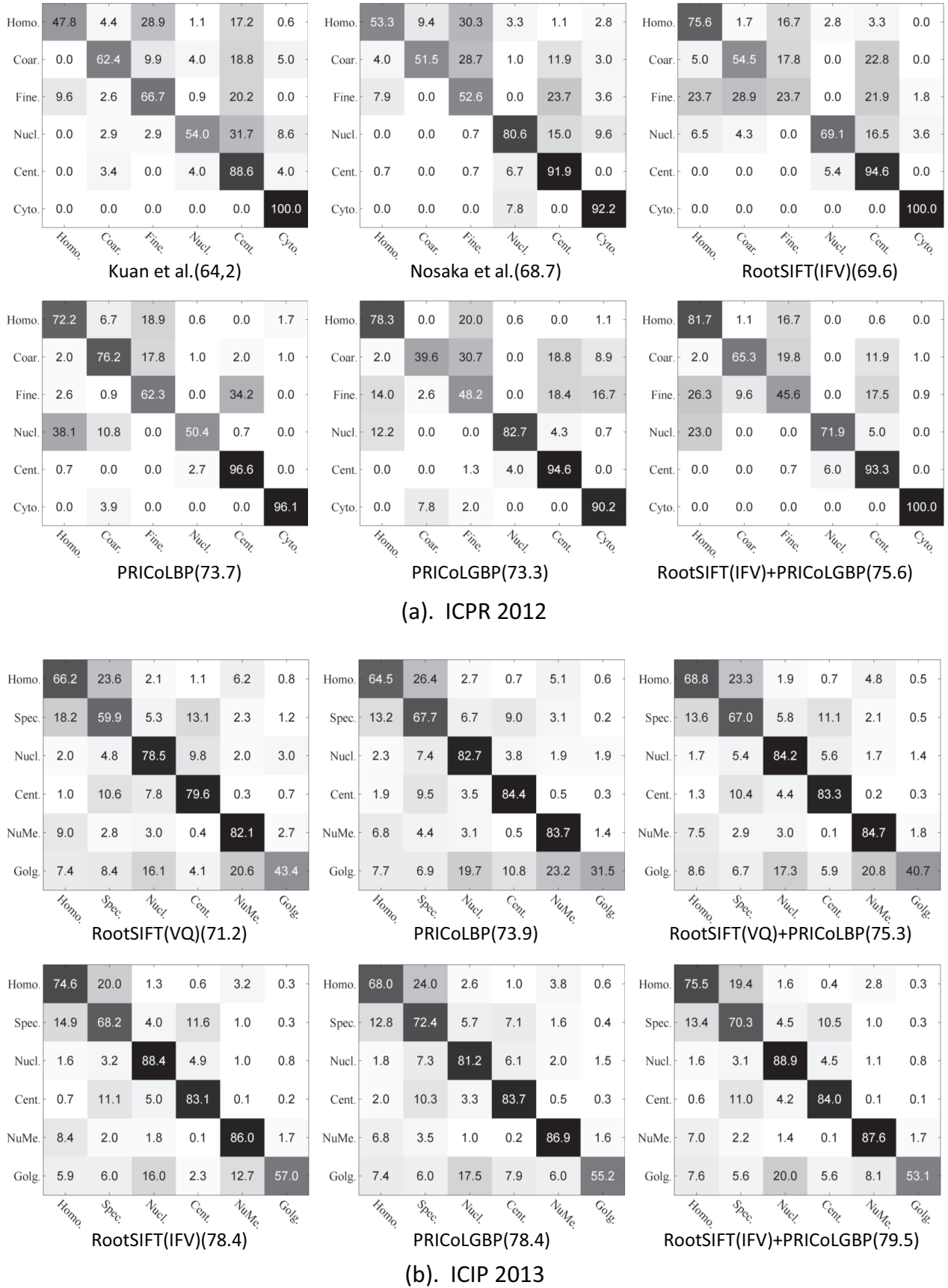}%{figures/examples.eps}
\end{center}
   \caption{Classification confusion matrixs and averaged accuracies on (a). ICPR 2012 and (b) ICIP 2013 contest data sets.}
\label{fig:icprconfusion}
\end{figure*}
\end{multicols}

%
%\begin{multicols}{2}
%\begin{figure*}[t]
%\begin{center}
%\small
% \includegraphics[width=1.1\linewidth]{figures/icipnew.eps}%{figures/examples.eps}
%\end{center}
%   \caption{Classification confusion matrix and averaged accuracies on ICIP 2013 contest data set.}
%\label{fig:icipconfusion}
%\end{figure*}
%\end{multicols}

\subsection{Comparison with the State-of-the-art Methods}
%For all experiments, we set \textcolor{blue}{the soft margin parameter ``C = 1''} without tuning any parameter.

{\bf{Experiments on ICPR 2012 contest.}} For this dataset, we evaluate seven methods, including PRICoLBP, PRICoLGBP, RootSIFT(IFV), the combination of PRICoLGBP and RootSIFT(IFV), and the top three methods in ICPR 2012 contest. For PRICoLBP and PRICoLGBP, we use the green channel. For RootSIFT(IFV), we use the gray image. In IFV, since the sampled patch is $41\times 41$, when the minimal size of the image is less than 50, we will resize the image to the minimal size 64 while keeping the ratio between the height and width. Since the dataset is very small, for the PRICoLBP and PRICoLGBP, we directly use SVM with ${\chi}^2$ kernel. For the RootSIFT(IFV), and the combination of PRICoLGBP and RootSIFT(IFV), we use linear SVM. The classification confusion matrix and averaged accuracies using the provided experimental setup by the ICPR 2012 contest organizers are shown in Fig. \ref{fig:icprconfusion}(a).

We have the following observations from Fig. \ref{fig:icprconfusion}(a):
\begin{itemize}
\item Texture based methods works better than the shape based methods. For specimen, PRICoLBP achieves 73.7\% which is higher than RootSIFT(IFV) (69.6\%).

\item IFV encoding with RootSIFT works well on this dataset, and slightly outperforms winner of ICPR 2012 contest.

\item The combination of our PRICoLGBP and RootSIFT(IFV) significantly outperforms the winner of ICPR 2012, and performs better than the latter on four categories including ``Homogeneous'', ``Coarse Speckled'', ``Centromere'' and ``Cytoplasmic'', and worse on the categories ``Nucleolar'' and ``Fine Speckled''.

\end{itemize}

%%%
%%%\begin{table}[h]
%%%\caption{Comparison with State-of-the-art methods on ICPR 2012 contest dataset.}              % title of Table
%%%\centering                                     % used for centering table
%%%%\large
%%%\normalsize
%%%%\small
%%%\begin{tabular}{|c | c |  }
%%%\hline
%%%
%%%Methods               &   Accuracy          \\[0.1cm]
%%%\hline
%%%Kuan et al.           &     64.2            \\[0.1cm]
%%%\hline
%%%Xiangfei et al.       &     65.8            \\[0.1cm]
%%%\hline
%%%Nokasa et al.         &     68.7            \\[0.1cm]
%%%\hline
%%%RootSIFT(IFV)             &     69.6            \\[0.1cm]
%%%\hline
%%%%PRICoLBP (Gray)       &     72.5            \\[0.1cm]
%%%%\hline
%%%PRICoLBP           &     73.7            \\[0.1cm]
%%%\hline
%%%%Gabor PRICoLBP (Gray) &     73.7            \\[0.1cm]
%%%%\hline
%%%Gabor-PRICoLBP     &     73.3            \\[0.1cm]
%%%%\hline
%%%%SIFT(FV)+Gabor PRICoLBP (R)             &                 \\[0.1cm]
%%%\hline
%%%Gabor-PRICoLBP + RootSIFT(IFV)  &  \bf{75.6} \\[0.1cm]
%%%\hline
%%%\end{tabular}
%%%\label{table:icprstate}                   % is used to refer this table in the text
%%%\end{table}
%%%
%%%
%
%The confusion matrix in Fig. \ref{fig:icprconfusion} shows the combination of our Gabor-PRICoLBP and RootSIFT(IFV) performs better than the winner of ICPR 2012 contst on five categories including ``Homogeneous'', ``Coarse Speckled'', ``Nucleolar'', ``Centromere'' and ``Cytoplasmic'', and worse on the category ``Fine Speckled''.

It should be noted that the experimental results on ICPR contest dataset are sensitive to the classifier's parameter C. We used the training set to conduct cross-validation to get a good C. Since the number of all specimens in ICPR 2012 contest is limited, thus, we use leave-one-out strategy to make cross-validation.
%
%
%To avoid overfitting, we just set the C = 1 for all experiments although we also found that when C = 1.5, PRICoLBP on Green channels can achieve 77.1\%.
%According to our experiments, we find that the scale of ICPR 2012 contest data set is small and is not enough to evaluate different algorithms. The scale of ICIP 2013 data is about 10 times of ICPR 2012 data set. We found that the classification performance on ICIP 2013 data set are not sensitive to classifier's parameter C.

{\bf{Experiments on ICIP 2013 contest.}} We evaluate and compare six methods including RootSIFT(VQ), PRICoLBP, the combination of RootSIFT with VQ and PRICoLBP, PRICoLGBP, RootSIFT(IFV), and the combination of PRICoLGBP and RootSIFT(IFV).
Here, we use the experimental setup ``D''. The features are all normalized, and a linear SVM. The classification confusion matrix and averaged accuracies based on 10 random repeats are shown in Fig. \ref{fig:icprconfusion}(b).

%with the soft margin parameter C = 40 is used

The confusion matrix in Fig. \ref{fig:icprconfusion}(b) indicates that:
\begin{itemize}

\item Multi-resolution PRICoLGBP texture feature significantly outperforms the single-resolution PRICoLBP, and improves the performance from 73.9\% to 78.4\%. PRICoLGBP significantly improves the PRICoLBP on several categories such as ``Speckled'' and ``Golgi'', and has high performance on other categories. Compared with RootSIFT(VQ), RootSIFT(IFV) significantly outperforms the former on all categories. This fully demonstrates the effectiveness of IFV encoding methods.

%\item IFV encoding sharply outperforms the VQ encoding, and improves the latter from 71.2\% to 78.4\%.

\item The combination between texture and shape features outperforms each of them. For specimen, the combination of RootSIFT(VQ) and PRICoLBP improves the PRICoLBP (73.9\%) and RootSIFT(VQ) (71.2\%) to 75.3\%. And, the combination of PRICoLGBP and RootSIFT(IFV) greatly improves the Shen's method (the winner of ICIP 2013) from 75.3\% to 79.5\%.

%\item The combination of PRICoLGBP and Bag of RootSIFT (IFV) improves the combination of PRICoLBP and RootSIFT (VQ) on all categories especially on the categories ``Homogeneous'' (6.8\% improvement) and ``Golgi'' (12.6\% improvement).

\item The category ``Golgi'' obtains the lowest performance, this is due to the less training sampling in this category. The most confusing pairs are ``Golgi'' and ``Nucleolar'', and ``Speckled'' and ``Homogeneous''. It is easy to find that from Fig. \ref{fig:icip}, the shape and texture structures in ``Homogeneous'' and ``Speckled'' look similar.

\end{itemize}

{\bf{Comparision with the Winner of ICPR 2014 contest.}}
Recently, the results of ICPR 2014 HEp-2 Cell classification have been released. In this part, we compare our method with the winner of ICPR 2014 contest \cite{manivannan2014hep}. We strictly follow the experimental protocol of the winner, and use the leave-one-specimen-out strategy. According to the specimen IDs, we can split the data into training and validation sets. Since we have 83 different specimens, in each test, we use 82 specimens for training and the left one for testing. The result of the ICPR 2014 winner is shown in Table \ref{table:icpr2014winner}(a), and our result based on PRIGCoLBP and RootSIFT(IFV) is shown in Table \ref{table:icpr2014winner}(b).

%
%\begin{table}[h]
%\caption{Confusion matrix of our method for leave-one-specimen-out experiments on I3A Task-1 dataset.}       % title of Table
%\centering                                     % used for centering table
%%\large
%\normalsize
%%\small
%\begin{tabular}{ c  c c c c c c  }
%\hline
%
%               &   Homo.  &  Spec.  &  Nucl.  &  Cent.  &  NuMe.  & Golgi \\[0.1cm]
%\hline
%Homo.          &   \bf{81.32}   &  15.56  &  0.80   &  0.12   &  1.96   &  0.24 \\[0.1cm]
%Spec.          &   12.19   &  \bf{73.86}  &  3.74   &  9.29   &  0.78   &  0.14 \\[0.1cm]
%Nucl.          &   1.58   &  2.23   &  \bf{92.49}  &  2.19   &  0.92   &  0.58 \\[0.1cm]
%Cent.          &   0.15   &  10.98  &  3.43   &  \bf{85.41}  &  0   &  0.04 \\[0.1cm]
%NuMe.          &   6.52   &  1.54   &  1.27   &  0.05   &  \bf{88.04}  &  2.58 \\[0.1cm]
%Colgi.         &   8.56   &  4.70   &  19.34  &  3.31   &  4.97   &  \bf{59.12}\\[0.1cm]
%\hline
%
%
%\hline
%\end{tabular}
%\label{table:icpr2014our}                   % is used to refer this table in the text
%\end{table}

According to Table \ref{table:icpr2014winner}, the Mean Class Accuracy (MCA) for the winner of ICPR 2014 contest is 80.25\%, and our method achieves a comparable performance 80.04\%. It should be noted that Manivannan et al. used four types of features, and for each feature, they trained 4 models, and the final classification result is based on summation of probability of 16 classifier. However, we only trained one model using the combination of the RootSIFT (IFV) and PRICoLGBP features. We can find that our method performs better on ``Nucleolar'' and ``Golgi'', and performs worse on ``Homogeneous'', ``Speckled'' and ``Centromere''.

\begin{table}[h]
\caption{Confusion matrixs of (a). the ICPR 2014 Contest Winner, and (b). Our method for leave-one-specimen-out experiments on I3A Task-1 dataset.}       % title of Table
\centering                                     % used for centering table
%\large
\normalsize
%\small
\begin{tabular}{ c  c c c c c c  }

\multicolumn{7}{c}{(a). The winner of ICPR 2014 contest.} \\
\hline
               &   Homo.  &  Spec.  &  Nucl.  &  Cent.  &  NuMe.  & Golgi \\[0.1cm]
\hline
Homo.          &   \bf{81.8}   &  15.00  &  0.76   &  0.20   &  2.04   &  0.20 \\[0.1cm]
Spec.          &   8.87   &  \bf{77.36}  &  3.67   &  9.18   &  0.74   &  0.18 \\[0.1cm]
Nucl.          &   1.12   &  3.89   &  \bf{90.65}  &  2.08   &  1.27   &  1.00 \\[0.1cm]
Cent.          &   0.47   &  10.87  &  2.85   &  \bf{85.66}  &  0.04   &  0.11 \\[0.1cm]
NuMe.          &   6.30   &  2.04   &  1.40   &  0.27   &  \bf{88.04}  &  1.95 \\[0.1cm]
Colgi.         &   5.66   &  3.73   &  20.72  &  2.35   &  9.53   &  \bf{58.01}\\[0.1cm]
\hline

\\
\multicolumn{7}{c}{(b). Our method.} \\
\hline
               &   Homo.  &  Spec.  &  Nucl.  &  Cent.  &  NuMe.  & Golgi \\[0.1cm]
\hline
Homo.          &   \bf{81.32}   &  15.56  &  0.80   &  0.12   &  1.96   &  0.24 \\[0.1cm]
Spec.          &   12.19   &  \bf{73.86}  &  3.74   &  9.29   &  0.78   &  0.14 \\[0.1cm]
Nucl.          &   1.58   &  2.23   &  \bf{92.49}  &  2.19   &  0.92   &  0.58 \\[0.1cm]
Cent.          &   0.15   &  10.98  &  3.43   &  \bf{85.41}  &  0   &  0.04 \\[0.1cm]
NuMe.          &   6.52   &  1.54   &  1.27   &  0.05   &  \bf{88.04}  &  2.58 \\[0.1cm]
Colgi.         &   8.56   &  4.70   &  19.34  &  3.31   &  4.97   &  \bf{59.12}\\[0.1cm]

\hline
\end{tabular}
\label{table:icpr2014winner}                   % is used to refer this table in the text
\end{table}

\section{Conclusion}
\label{sec:conclusion}
We have proposed an efficient and effective method for automatic classification of HEp-2 cell via using multiresolution texture and richer shape information.  Specifically, we proposed to capture the multi-resolution
texture information by a novel Pairwise Rotation Invariant Co-occurrence of Local Gabor Binary Pattern (PRICoLGBP) descriptor, depict the richer shape information by using an Improved Fisher Vector (IFV) model with RootSIFT features which are sampled from large image patches in multiple scales, and combine them properly.  We have systematically evaluated the proposed approach on the ICPR 2012, ICIP 2013, and ICPR 2014 contest data sets. The proposed approach significantly outperformed the winners of ICPR 2012 and ICIP 2013 contests, and yileded comparable performance with the winner of the newly released ICPR 2014 contest.

%
%We evaluate the proposed features on the ICPR 2012 and ICIP 2013 contest data sets. The proposed method significantly outperforms the winner of ICPR 2012 and ICIP 2013 contest. Specifically, on ICPR contest, our method outperforms the ICPR winner by 7\%, and on ICIP 2013 contest dataset, our method outperforms the winner by more than 4\%. It should be noted that the difference between the $1$-st and $2$-nd places in ICIP 2013 contest is only 0.1\%.
%%
%%The proposed texture and shape features have many potential applications. They are not limited to the IIF HEp-2 cell classification. We believe that the proposed methods can generalize to many medical applications, such as Tissue Classification \cite{song2013feature} \cite{van1999automated} and Breast
%%Sonographic Tumor Diagnosis \cite{yang2013robust} and etc.

\section{Acknowledgment}
The authors would like to thank Andrea Vedaldi for sharing the Vlfeat toolbox, and Rong-En Fan et al. for sharing the Liblinear toolbox. We also want to thank the organizer of ICPR 2012 and ICIP 2013 HEp-2 Cell classification contests for releasing the HEp-2 cell data sets. These two contests substantially put forward the research on HEp-2 cell classification. This work was supported by the Academy of Finland and Infotech Oulu.

{
\large
\bibliographystyle{IEEEtran}
\bibliography{egbib}
}

\end{document}